\documentclass[10pt,twocolumn,letterpaper]{article}

\usepackage{cvpr}
\usepackage{times}
\usepackage{epsfig}
\usepackage{booktabs}
\usepackage{graphicx}
\usepackage{amsmath}
\usepackage{amssymb}
\usepackage{xcolor}
\usepackage{multirow}
\usepackage{caption}
\usepackage[symbol]{footmisc}

\usepackage[pagebackref=true,breaklinks=true,letterpaper=true,colorlinks,bookmarks=false]{hyperref}

\cvprfinalcopy 

\def\cvprPaperID{2444} 

\ifcvprfinal\fi
\begin{document}

\title{DeepLight: Learning Illumination for Unconstrained Mobile Mixed Reality} 

\author{\vspace{2pt} Chloe LeGendre\textsuperscript{1,2 \footnotemark[1]} \hspace{20pt} Wan-Chun Ma\textsuperscript{1} \hspace{20pt} Graham Fyffe\textsuperscript{1} \hspace{20pt} John Flynn\textsuperscript{1}\\ 
\vspace{2pt} Laurent Charbonnel\textsuperscript{1} \hspace{20pt} Jay Busch\textsuperscript{1} \hspace{20pt} Paul Debevec\textsuperscript{1}\\
\textsuperscript{1}Google Inc.  \hspace{20pt} \textsuperscript{2}USC Institute for Creative Technologies\vspace{-10pt}\\
}

\renewcommand{\thefootnote}{\fnsymbol{footnote}}


\twocolumn[{%
\renewcommand\twocolumn[1][]{#1}%
\maketitle
\thispagestyle{empty}
\begin{center}
    \centering
    \footnotesize
        \vspace{-12pt}
		\begin{tabular}{c@{\quad \enskip}c@{ }c@{ }c@{ }c@{ }c@{ }c@{ }} 
			\includegraphics[height=1.6in]{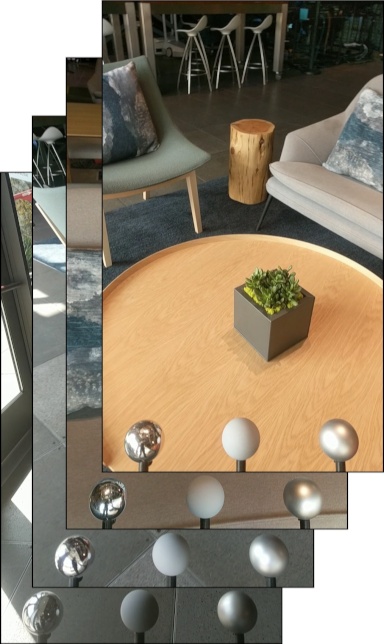} &
			\includegraphics[height=1.6in]{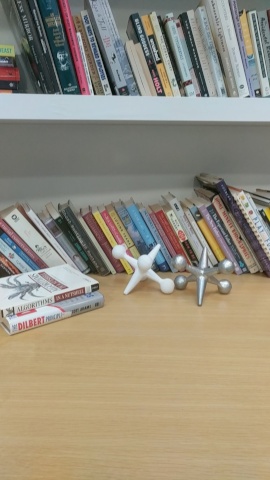} &
			\includegraphics[height=1.6in]{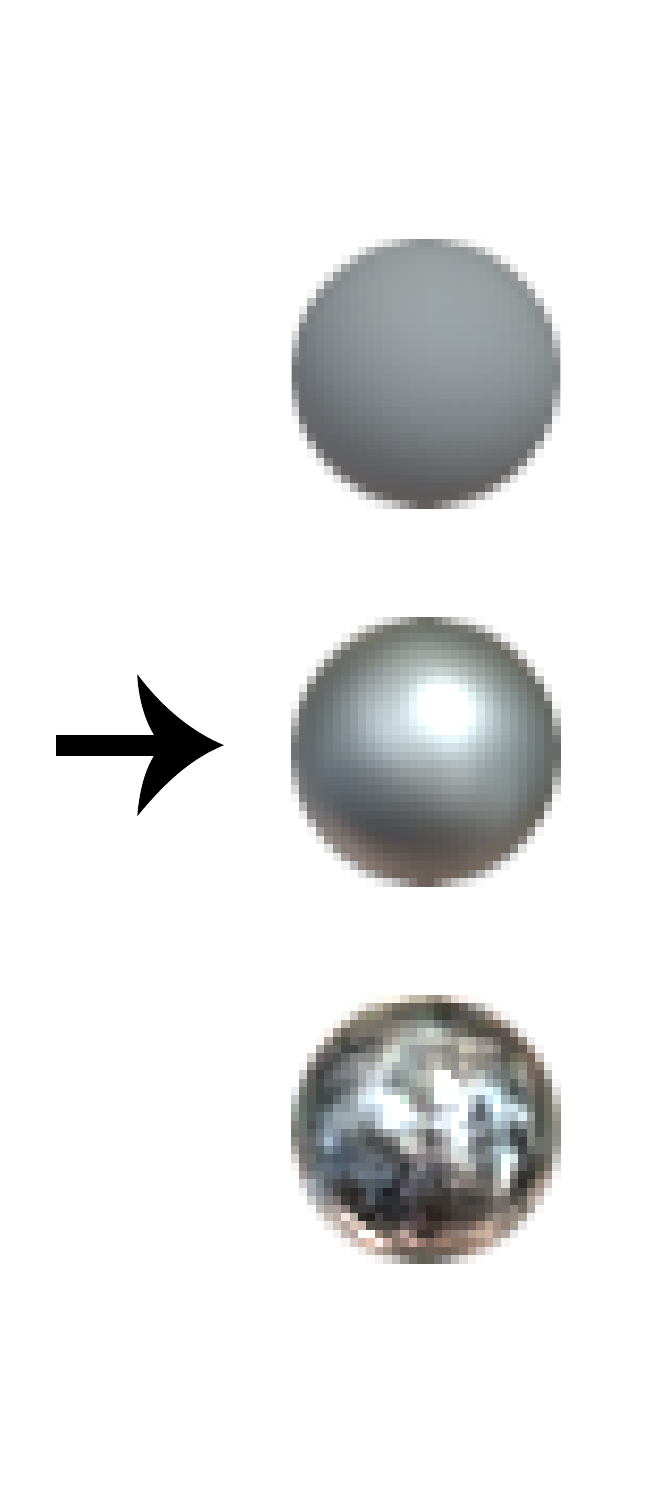} &
			\includegraphics[height=1.6in]{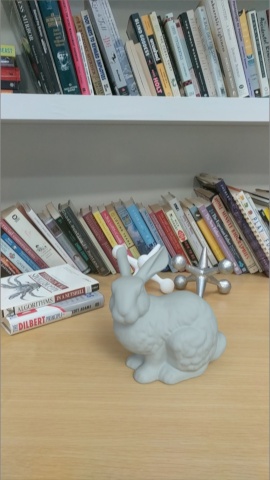} &
			\includegraphics[height=1.6in]{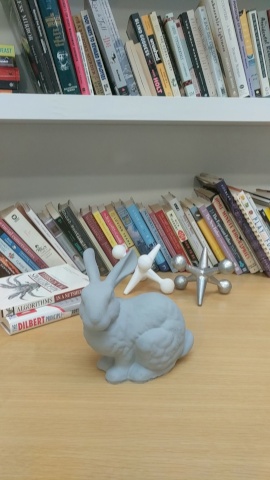} &
			\includegraphics[height=1.6in]{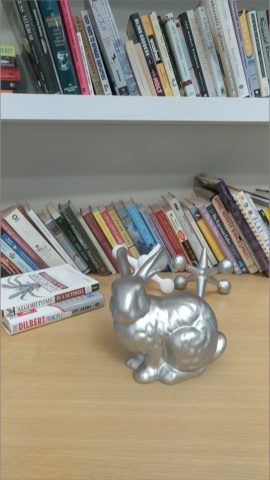} &
			\includegraphics[height=1.6in]{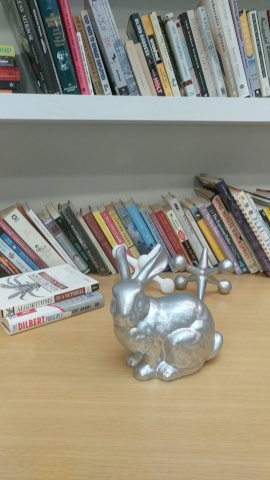} \\
			(a) training data & (b) input image & (c) output lighting & (d) rendered object & (e) real object & (f) rendered object & (g) real object\\
		\end{tabular}
	\vspace{-8pt}
    \captionof{figure}{\small Given an arbitrary low dynamic range (LDR) input image captured with a mobile device (b), our method produces omnidirectional high dynamic range lighting (c, lower) useful for rendering and compositing virtual objects into the scene. We train a CNN with LDR images (a) containing three reflective spheres, each revealing different lighting cues in a single exposure. (d) and (f) show renderings produced using our lighting, closely matching photographs of real 3D printed and painted objects in the same scene (e, g).}
\end{center}%
}]

\begin{abstract} \vspace{-10pt}
 We present a learning-based method to infer plausible high dynamic range (HDR), omnidirectional illumination given an unconstrained, low dynamic range (LDR) image from a mobile phone camera with a limited field of view (FOV). For training data, we collect videos of various reflective spheres placed within the camera's FOV, leaving most of the background unoccluded, leveraging that materials with diverse reflectance functions reveal different lighting cues in a single exposure. We train a deep neural network to regress from the LDR background image to HDR lighting by matching the LDR ground truth sphere images to those rendered with the predicted illumination using image-based relighting, which is differentiable. Our inference runs at interactive frame rates on a mobile device, enabling realistic rendering of virtual objects into real scenes for mobile mixed reality. Training on automatically exposed and white-balanced videos, we improve the realism of rendered objects compared to the state-of-the art methods for both indoor and outdoor scenes. \footnotetext[1]{Work completed while interning at Google.}
 \vspace{-12pt}
\end{abstract}

\renewcommand*{\thefootnote}{\arabic{footnote}}
\vspace{-5pt}
\section{Introduction}
Compositing rendered virtual objects into photographs or videos is a fundamental technique in mixed reality, visual effects, and film production. The realism of a composite depends on both geometric and lighting related factors. An object ``floating in space" rather than placed on a surface will immediately appear fake; similarly, a rendered object that is too bright, too dark, or lit from a direction inconsistent with other objects in the scene can be just as unconvincing. In this work, we propose a method to estimate plausible illumination from mobile phone images or video to convincingly light synthetic 3D objects for real-time compositing.

\begin{figure}[ht]
    \vspace{-5pt}
	\centerline{
		\begin{tabular}{c@{ }c@{ }} 
			\includegraphics[height=1in]{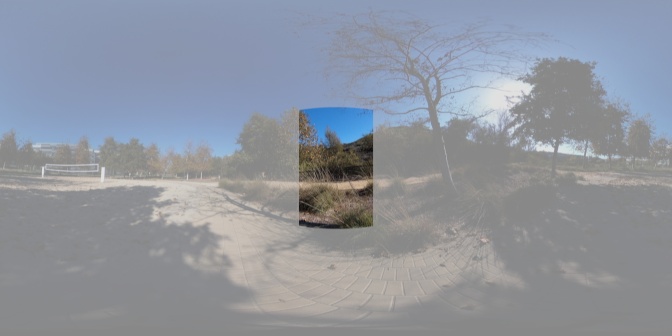} &
			\includegraphics[height=1in]{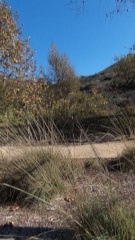} \\
		\end{tabular}}
		\vspace{-5pt}
		\caption{{The field of view (FOV) of mobile phone video (inset shown in full color), relative to the 360$^{\circ}$ environment.}}
		\vspace{-5pt}
		\label{fig:FOV}
\end{figure}

Estimating scene illumination from a single photograph with low dynamic range (LDR) and a limited field of view (FOV) is a challenging, under-constrained problem. One reason is that an object's appearance in an image is the result of the light arriving from the full sphere of directions around the object, including from directions outside the camera's FOV.  However, in a typical mobile phone video, only 6\% of the panoramic scene is observed by the camera (see Fig. \ref{fig:FOV}). Furthermore, even light sources within the FOV will likely be too bright to be measured properly in a single exposure if the rest of the scene is well-exposed, saturating the image sensor due to limited dynamic range and thus yielding an incomplete record of relative scene radiance. To measure this missing information, Debevec \cite{debevec:1998:rendering} merged omnidirectional photographs captured with different exposure times and lit synthetic objects with these high dynamic range (HDR) panoramas using global illumination rendering.  But in the absence of such measurements, professional lighting artists often create convincing illumination by reasoning on cues like shading, geometry, and context, suggesting that a background image alone may provide sufficient information for plausible lighting estimation.

As with other challenging visual reasoning tasks, convolutional neural networks (CNNs) comprise the state-of-the-art techniques for lighting estimation from a limited-FOV, LDR image, for both indoor \cite{Gardner:2017:Indoor} and outdoor \cite{hold:2017:deep} scenes. Na\"ively, many pairs of background images and lighting (HDR panoramas) would be required for training; however, capturing HDR panoramas is complex and time-consuming, so no such dataset exists for both scene types. For indoor scenes, Gardner et al. \cite{Gardner:2017:Indoor} first trained a network with many LDR panoramas \cite{xiao:2012:sun360}, and then fine-tuned it with 2100 captured HDR panoramas. For outdoor scenes, Hold-Geoffroy et al. \cite{hold:2017:deep} fit a sky model to LDR panoramas for training data. We also use a CNN, but our model generalizes to both indoor and outdoor scenes and requires no HDR imagery. 

In this work, our training data is captured as LDR images with three spheres held within the bottom portion of the camera's FOV (Fig.\ \ref{fig:multi_prober}), each with a different material that reveals different cues about the scene's ground truth illumination. For instance, a mirrored sphere reflects omnidirectional, high-frequency lighting, but, in a single exposure, bright light source reflections usually saturate the sensor so their intensity and color are misrepresented. A diffuse gray sphere, in contrast, reflects blurred, low-frequency lighting, but captures a relatively complete record of the total light in the scene and its general directionality. We regress from the portion of the image unoccluded by the spheres to the HDR lighting, training the network by minimizing the difference between the LDR ground truth sphere images and their appearances \textit{rendered} with the estimated lighting. We first measure each sphere's reflectance field as in \cite{debevec:2000:acquiring}. Then, during training, we render the spheres with the estimated HDR lighting using image-based relighting \cite{debevec:2000:acquiring,nimeroff:1995:efficient}, which is differentiable. Furthermore, we add an adversarial loss term to improve recovery of plausible high-frequency illumination. As only one exposure comprises each training example, we can capture \textit{videos} of real-world scenes, which increases the volume of training data and gives a prior on the automatic exposure and white-balance of the camera.

For a public benchmark, we collect 200k new images in indoor and outdoor scenes, each containing the three different reflective spheres. We show on a random subset that our method out-performs the state-of-the-art lighting estimation techniques for both indoor and outdoor scenes for mobile phone imagery, as our inferred lighting more accurately renders synthetic objects. Furthermore, our network runs at interactive frames rates on a mobile device, and, when used in combination with real-time rendering techniques, enables more realistic mobile mixed reality composites. 

In summary, our key contributions are:
\begin{itemize} \vspace{-5pt}
\item A data collection technique and dataset of paired lighting reference spheres and background images (200k examples) for training a lighting estimation algorithm.
\item A CNN-based method to predict plausible omnidirectional HDR illumination from a single unconstrained image. To the best of our knowledge, ours is the first to generalize to both indoor and outdoor scenes. 
\item A novel image-based relighting rendering loss function, used for training the HDR lighting inference network using \textit{only} LDR data.
\end{itemize}

\section{Related work}
Debevec \cite{debevec:1998:rendering} rendered synthetic objects into photographs of real-world scenes using HDR panoramas as lighting. These can be captured by photographing a mirrored sphere or stitching together wide-angle views using multiple exposures \cite{debevec:1998:rendering, stumpfel:2004:direct}. Recording HDR video of a mirror ball \cite{Waese:2002:ART, Unger:2006:DSL} has been used for real-time capture of image-based lighting environments. Our goal is to estimate HDR lighting given only a single LDR image with a limited, but fixed, FOV. Key to our technique is that spheres with diverse reflectance functions (BRDFs) reveal different lighting cues, enabling us to record training data using a standard LDR video stream. This has been previously leveraged for sun intensity recovery from clipped panoramas using a diffuse, gray sphere \cite{reinhard:2010:high, debevec:2012:single}.

The appearance of a scene depends on its geometry, reflectance properties, and lighting, as well as the camera's exposure, color balance, and depth-of-field. The joint recovery of geometry, reflectance, and lighting, known as the inverse rendering problem, has been a core computer vision challenge \cite{yu:1999:inverse, ramamoorthi:2001:signal}. Intrinsic image decomposition \cite{barrow:1978:recovering} separates an image into shading and reflection; however, shading is an \textit{effect} of lighting, not its direct observation. While recent approaches jointly inferred material reflectance and illumination from an object comprised of an unknown material \cite{Meka:2018:Lime, liu:2017:material}, one or more images of a segmented object \cite{Wang:2018:joint, lopez:2013:multiple}, specular objects of a known class \cite{rematas:2016:deep, georgoulis:2016:delight}, or with measured or known geometry \cite{lombardi:2016:reflectance, weber:2018:objects, mandl:2017:learning, gruber:2012:real, barron:2013:intrinsic}, we estimate lighting from unconstrained images with unknown geometry and arbitrarily complex scenes.

Khan et al. \cite{Khan:2006:image} projected a limited-FOV HDR image onto a hemisphere and flipped it to infer 360$^{\circ}$ lighting. For LDR images, Karsch et al. \cite{karsch:2014:automatic} estimated a scene's geometry and diffuse albedo, detected in-view light sources, and, for unseen lights, found a matching LDR panorama from a database \cite{xiao:2012:sun360}. They promoted the result to HDR, minimizing a diffuse scene rendering loss. For indoor scenes, Gardner et al. \cite{Gardner:2017:Indoor} learned a mapping from a limited FOV LDR image to HDR lighting using a CNN. Noting the lack of HDR panoramas, they leveraged the same LDR panorama dataset \cite{xiao:2012:sun360} to regress first from the input image to a LDR panorama and light source locations and then refined the model for light source intensities with 2100 new, captured HDR panoramas. Though demonstrating state-of-the-art results, they noted two key limitations. First, the predicted LDR panorama and HDR light sources were white-balanced to match the input image using the Gray World assumption \cite{Buchsbaum:1980:ASP}. Second, renderings improved when an artist manually tuned the predicted lighting intensity. We propose a novel rendering-based loss function that allows our network to learn both the colors and intensities of the incident illumination relative to the input image, without HDR imagery. Furthermore, we propose a lighting model that generalizes to both indoor and outdoor scenes, though outdoor HDR lighting estimation from a single image or from a LDR panorama has also received attention, as the sun and sky afford lower dimensional lighting parameterizations \cite{lalonde:2009:estimating, lalonde:2014:collections, hold:2017:deep, zhang:2017:outdoor}. Cheng et al. \cite{cheng:2018:learning} estimated lighting from opposing views within a panorama for indoor and outdoor scenes, but did not consider single image inputs.

Several recent works estimate lighting from faces, modeling image formation via rendering within the ``decoder" of an encoder-decoder architecture \cite{calian:2018:faces, tewari:2017:mofa, tewari:2018:FaceModel, zhou:2018:label, sengupta:2018:sfsnet, shu:2017:neural}. However, all have relied on simple or low frequency shading models. In contrast, we render objects during training using image-based relighting (IBRL) \cite{debevec:2000:acquiring, nimeroff:1995:efficient}, forming new images as a linear combination of reflectance basis images, avoiding an analytic shading model altogether. Xu et al. \cite{xu:2018:deep} trained a network to perform IBRL, jointly \textit{learning} a low-dimensional reflectance basis and renderer, rather than applying IBRL as a fixed function as we do. Hold-Geoffroy et al. \cite{hold:2017:deep} and Cheng et al. \cite{cheng:2018:learning} used a synthetic Lambertian reflectance basis in a rendering loss term but did not use a photographed basis or consider multiple BRDFs.

\section{Method}
Here we describe how we acquire our training data, our network architecture, and the loss functions of our end-to-end lighting estimation method.

\subsection{Training Data Acquisition and Processing}

\begin{figure}[h]
	\centerline{
		\begin{tabular}{c@{ }c@{ }c@{ }} 
			\includegraphics[width=1.7in]{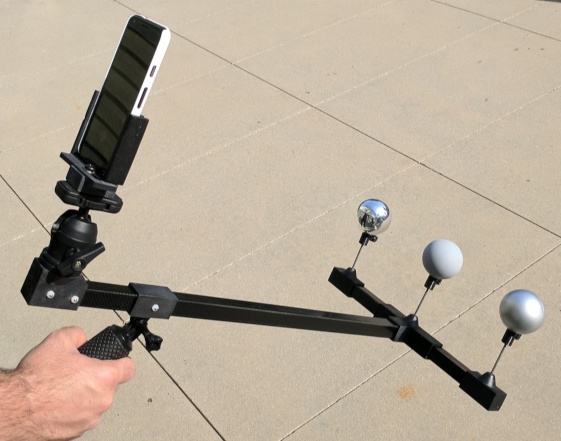} &
			\includegraphics[width=0.75in]{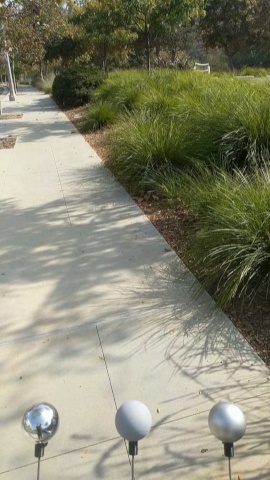} &
			\includegraphics[width=0.75in]{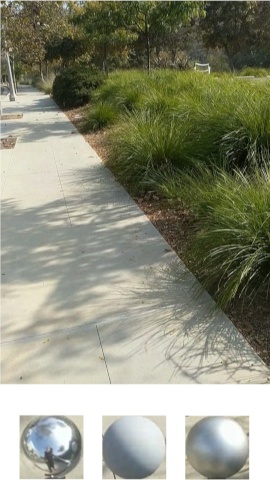} \\
		\end{tabular}}
		\vspace{-5pt}
		\caption{{Left: Capture apparatus. Center: Example frame. Right: Processed data (top: input; bottom: ground truth).}}
		\vspace{-10pt}
		\label{fig:multi_prober}
\end{figure}

Gardner et al. \cite{Gardner:2017:Indoor} fine-tuned a pre-trained network using 2100 HDR panoramas, fewer examples than would be typically required for deep learning without pre-training. However, our key insight is that we can infer HDR lighting from \textit{only} LDR images with reference objects in the scene, provided they span a range of BRDFs that reveal different lighting cues. Thus, we collect LDR images of indoor and outdoor scenes, where each contains three spheres located in the bottom portion of the camera's FOV, occluding as little of the background as possible (Fig.\ \ref{fig:multi_prober}, center). The three spheres are plastic holiday ornaments with diverse finishes that differently modulate the incident illumination: mirrored silver, matte silver (rough specular), and diffuse gray (spray-painted), measured as 82.7\%, 64.4\%, and 34.5\% reflective respectively. We built a capture rig to fix the sphere-to-phone distance, stabilizing the sphere positions in each image (see Fig.\ \ref{fig:multi_prober}, left).

As we require only LDR input imagery, we collect portrait HD ($1080 \times 1920$) \textit{video} at 30 fps, rather than static photographs. This increases the speed of training data acquisition compared with HDR panoramic photography, enabling the capture of millions of images, albeit with significant redundancy for adjacent frames. The videos feature automatic exposure and white balance, providing a prior to help disambiguate color, reflectance, and illumination.

We locate the three spheres in each video frame by detecting circular boundaries in the optical flow field between neighboring frames (see supplemental materials for more details), though marker-based tracking could also be used. We re-sample cropped images of the spheres using an idealized camera model oriented towards the sphere center with a view frustum tangent to the sphere on all four sides to eliminate perspective distortion. For the background images, we remove the lower 20\% of each frame during both training and inference. The final training data consists of cropped background images, each paired with a set of three cropped spheres, one per BRDF (Fig.\ \ref{fig:multi_prober}, right).

\begin{figure*}[ht]
    \centering
    \includegraphics[width=6.7in]{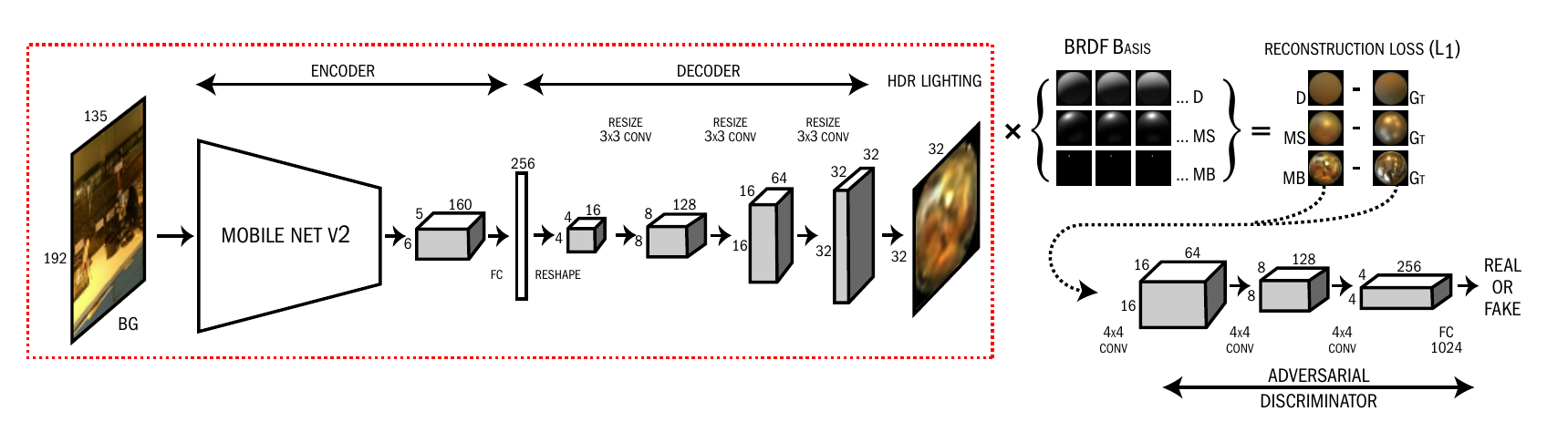}
    \vspace{-8pt}
    \caption{\small{Overview of our network. We regress to HDR lighting from an LDR, limited-FOV input image captured with a mobile device. We include a multi-BRDF image-based relighting reconstruction loss for a diffuse(D), matte silver(MS), and mirror ball(MB) and an adversarial loss for the mirror ball only. Only the part outlined in red occurs at inference time.}}
    \label{fig:network}
    \vspace{-10pt}
\end{figure*}

\subsection{Network Architecture}

The input to the model is an unconstrained LDR, gamma-encoded image captured with a mobile phone, resized from the native cropped resolution of $1080 \times 1536$ to $135 \times 192$ and normalized to the range of $[-0.5, 0.5]$. Our architecture is an encoder-decoder type, where the encoder includes fast depthwise-separable convolutions \cite{howard:2017:mobilenets}. We use the first 17 MobileNetV2 \cite{sandler:2018:inverted} layers, processing the output feature maps with a fully-connected (FC) layer to generate a latent vector of size 256. For the decoder, we reshape this vector and upsample thrice by a factor of two to generate a $32\times32$ color image of HDR lighting. We regress to natural log space illumination as the sun can be more than five orders of magnitude brighter than the sky \cite{stumpfel:2004:direct}. Although we experimented with fractionally-strided convolutions, bilinear upsampling with convolutions empirically improved our results. We train the network to produce omnidirectional lighting in the mirror ball mapping \cite{reinhard:2010:high}, where each pixel in image space represents an equal solid angular portion of a sphere for direction $(\theta,\phi)$. Thus, the corners of the output image are unused, but this mapping allows for equal consideration of all lighting directions in the loss function, if desired. For network details, see Fig.\ \ref{fig:network}.

\subsection{Reflectance Field Acquisition}
Debevec et al. \cite{debevec:2000:acquiring} introduced the 4D reflectance field $R(\theta,\phi, x, y)$ to denote the image of a subject with pixels $(x, y)$ as lit from any lighting direction $(\theta,\phi)$ and showed that taking the dot product of the reflectance field with an HDR illumination map relights the subject to appear as they would in that lighting. During training, we use this method to render spheres with the predicted HDR lighting. We photograph reflectance fields for the matte silver and diffuse gray spheres using a computer-controllable sphere of white LEDs \cite{ma:2007:rapid}, spaced $12^{\circ}$ apart at the equator. This produces an anti-aliased reflectance field for the diffuse and matte silver sphere; however, this LED spacing aliases the mirror BRDF. As we infer lighting in a mirror ball mapping, we instead construct the mirror ball basis as a set of $32\times32$ one-hot matrices of size $32\times32$, scaled by its measured reflectivity. We convert the lighting bases for the other BRDFs to the same geometric and relative radiometric space. The photographed bases are normalized based on the incident light source color and converted to the mirror ball mapping, accumulating energy from the photographs for each new lighting direction $i$ for the set of directions on the $32\times32$ mirror ball using a Phong lobe ($n=64$) and super-sampling with a $4\times4$ grid of directions on a sphere.

\subsection{Loss Function}

To train the lighting prediction network, we minimize an image-based relighting loss and add an adversarial loss to ensure inference of plausible high-frequency illumination.
\vspace{-3pt}
\paragraph{Image-based relighting rendering loss:} We train the network by minimizing the reconstruction loss between the ground truth sphere images $I$ and rendered spheres lit with the predicted HDR lighting. With the reflectance fields $R(\theta,\phi, x, y)$, pixel values for each sphere lit by each lighting direction $(\theta,\phi)$ of the $32\times32$ mirror ball, we can compute a linear image $\hat{I}$ of each sphere under a novel lighting environment $\hat{L}$ as a linear combination of its basis images. Slicing the reflectance field into individual pixels $R_{x,y}(\theta,\phi)$, we generate $\hat{I}_{x,y}$ with \eqref{Eqn:IBRL}, where $L_{i}(\theta,\phi)$ represents the color and intensity of light in the novel lighting environment for the direction $(\theta,\phi)$:
\vspace{-2pt}
\begin{equation}
\hat{I}_{x,y} = \sum_{\theta,\phi}R_{x,y}(\theta,\phi) L_{i}(\theta,\phi).
\label{Eqn:IBRL}
\end{equation}

The network outputs $Q$, a log space image of omnidirectional HDR lighting in the mirror ball mapping, with pixel values $Q_{i}(\theta,\phi)$. Thus we render each sphere with \eqref{Eqn:LogIBRL}:
\vspace{-2pt}
\begin{equation}
\hat{I}_{x,y} = \sum_{\theta,\phi}R_{x,y}(\theta,\phi) e^{Q_{i}(\theta,\phi)}.
\label{Eqn:LogIBRL}
\end{equation}

The ground truth sphere images $I$ are LDR, 8-bit, gamma-encoded images, possibly with clipped pixels. Accordingly, we clip the rendered sphere images with a differentiable soft-clipping function $\Lambda$, $n = 40$:
\vspace{-2pt}
\begin{equation}
\Lambda (p)= 1 - \tfrac{1}{n}\log\big(1 + e^{-n(p-1)}\big).
\label{Eqn:Softclip}
\end{equation}

We then gamma-encode the clipped linear renderings with $\gamma$, to match $I$. We mask out the pixels in the corners of each ball image with a binary mask $\hat M$, producing the masked $L_{1}$ reconstruction loss $L_\text{rec}$ for BRDFs $b=[0,1,2]$, where $\lambda_{b}$ represents an optional weight for each BRDF:
\begin{equation}
L_\text{rec} = \sum_{b=0}^{2}\lambda_{b}
\big\|\hat M\odot (\Lambda({\hat I_{b}})^{\tfrac{1}{\gamma}} - \Lambda(I_{b}))\big\|_{1}.
\label{Eqn:L1}
\end{equation}

\vspace{-10pt}
\paragraph{Adversarial loss:} 
Minimizing only $\mathbf{E}[L_\text{rec}]$ produces blurred, low-frequency illumination. While this might be acceptable for lighting diffuse objects, rendering shiny objects with realistic specular reflections requires higher frequency lighting. Recent works in image inpainting and synthesis \cite{pathak:2016:context, ledig:2017:photo, yu:2018:generative, iizuka:2017:globally, yang:2017:high, song:2018:im2pano3d} leverage Generative Adversarial Networks \cite{goodfellow:2014:generative} for increased image detail, adding an adversarial loss to promote multi-modal outputs rather than a blurred mean of the distribution. We train our network in a similar framework to render \textit{plausible} clipped mirror ball images, of which we have many real examples. This is perceptually motivated, as humans have difficulty reasoning about reflected light directions \cite{pont:2006:material, te:2005:comparison}, which digital artists leverage when environment mapping \cite{blinn:1976:texture} reflective objects with arbitrary images. Furthermore, real-world lighting is highly regular, statistically \cite{dror:2004:statistical}. 

Similar to Pathak et al. \cite{pathak:2016:context}, we use an auxiliary discriminator network $D$ with our base CNN as the generator $G$. During training, $G$ tries to trick $D$, producing clipped mirror ball images appearing as ``real" as possible. $D$ tries to discriminate between real and generated images. We condition $D$ on a few pixels from the original image surrounding the ball: we sample the four corners of the cropped ground truth mirror ball image, and bilinearly interpolate a $32\times32$ hallucinated background, as if the mirror ball were removed. We then softclip and composite both the ground truth and predicted mirror ball onto this ``clean plate" with alpha blending (yielding $I_{c}$, $\hat I_{c}$) providing $D$ with local color cues and ensuring that samples from both sets have the same perceptual discontinuity at the sphere boundary. Given input image $x$, $G$ learns a mapping to $Q$, $G:{x}\rightarrow Q$, used to render a mirror ball with \eqref{Eqn:LogIBRL}. The adversarial loss term, then, is:
\begin{multline}
L_\text{adv} = \log D(\Lambda(I_{c})) \\
+ \log(1-D( \Lambda(
\textstyle\sum_{\theta,\phi}R(\theta,\phi)e^{G(x;\theta,\phi)})^{\tfrac{1}{\gamma}}
)).
\label{Eqn:Ladv}
\end{multline}
\vspace{-10pt}

\paragraph{Joint objective:} 
The full objective is therefore:
\begin{equation}
G^{*} = \arg \min_{G} \max_{D} \, (1-\lambda_\text{rec})\mathbf{E}[L_\text{adv}] + \lambda_\text{rec} \mathbf{E}[L_\text{rec}].
\label{Eqn:L}
\end{equation}
\vspace{-5pt}
\subsection{Implementation Details}

We use TensorFlow \cite{tensorflow:2015:whitepaper} and train for 16 epochs using the ADAM \cite{kinga:2015:adam} optimizer with $\beta_{1}=0.9$, $\beta_{2}=0.999$, a learning rate of 0.00015 for $G$, and, as is common, one 100$\times$ lower for $D$, alternating between training $D$ and $G$. We set $\lambda_\text{rec}=0.999$, with $\lambda_{b}=0.2, 0.6, 0.2$ for the mirror, diffuse, and matte silver BRDFs respectively, and use $\gamma=2.2$, as the camera's video mode employs image-dependent tone-mapping. We use a batch size of 32 and batch normalization \cite{Ioffe:2015:batchnorm} for all layers but the last of $G$ and $D$. We use ReLU6 activations for $G$ and ELU \cite{clevert:2016:ELU} for $D$. For our mobile demo (supplemental materials), we use TFLite. For data augmentation, we horizontally flip the input and ground truth images. We found that data augmentation by modifying white balance and exposure did not improve results, perhaps since they simulated unlikely camera responses.

\vspace{-5pt} \paragraph{Datasets:}
We collected 37.6 hours of training video using a Google Pixel XL mobile phone, in a variety of indoor and outdoor locations, times of day, and weather conditions, generating 4.06 million training examples. We bias the data towards imagery of surfaces or ground planes where one might want to place a virtual AR object. For test data, we collected 116 new one-minute videos (211.7k frames) with the same camera and separated them into four sets: unseen indoor and outdoor (UI, UO) and seen indoor and outdoor (SI, SO). ``Unseen" test videos were recorded in new locations, while the ``seen" were \textit{new} videos recorded in previously-observed environments. We evaluate our method on the following videos: 28 UI (49.3k frames), 27 UO (49.7k frames), 27 SI (49.9k frames), and 34 SO (62.7k frames). Test data will be publicly released.

\section{Evaluation}

\subsection{Quantitative Results}
Accurate lighting estimates should correctly render objects with arbitrary materials, so we measure lighting accuracy first using $L_\text{rec}$, comparing with ground truth LDR spheres. We show the average per-pixel $L_{1}$ loss for each \textit{unseen} test dataset for each material and the per-pixel linear RGB angular error $\theta_\text{rgb}$ for the diffuse ball, a distance metric commonly used to evaluate white-balance algorithms (see Hordley and Finlayson \cite{hordley:2004:angular}), in Table \ref{Table:1} (top). (Minimizing $\theta_\text{rgb}$ during training did not improve results.) We show results for \textit{seen} test sets in supplemental material.

\begin{figure}[ht]
\centering
\scriptsize
\vspace{-10pt}
\begin{tabular}{c@{ }c@{ }c@{ }c@{ }c@{ }c@{ }c@{ }c@{ }}
& & \scriptsize{\textbf{L\textsubscript{1(all)}}}& & & & & \\
\scriptsize{input} & \scriptsize{gt}&
\scriptsize{\textbf{+ L\textsubscript{adv}}} &
\scriptsize L\textsubscript{1(all)} &
\scriptsize L\textsubscript{1(m)}&
\scriptsize L\textsubscript{1(s)}&
\scriptsize L\textsubscript{1(d)} &\\

&
\includegraphics[width=20px]{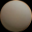} &
\includegraphics[width=20px]{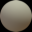}&  
\includegraphics[width=20px]{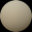}&   
\includegraphics[width=20px]{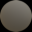}&  
\includegraphics[width=20px]{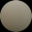}& 
\includegraphics[width=20px]{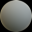}&   
(d)\\
\multirow[b]{-5.3}{*}{\includegraphics[height=65px]{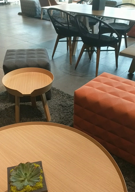}} &
\includegraphics[width=20px]{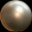} &
\includegraphics[width=20px]{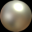}&  
\includegraphics[width=20px]{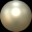}&   
\includegraphics[width=20px]{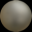}&  
\includegraphics[width=20px]{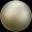}& 
\includegraphics[width=20px]{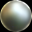}&   
(s)\\
&
\includegraphics[width=20px]{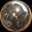} &
\includegraphics[width=20px]{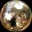}&  
\includegraphics[width=20px]{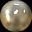}&   
\includegraphics[width=20px]{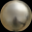}&  
\includegraphics[width=20px]{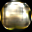}& 
\includegraphics[width=20px]{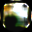}&   
(m)\\[1ex]

 &
\includegraphics[width=20px]{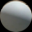} &
\includegraphics[width=20px]{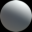}&  
\includegraphics[width=20px]{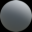}&   
\includegraphics[width=20px]{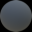}&  
\includegraphics[width=20px]{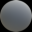}& 
\includegraphics[width=20px]{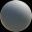}&   
(d)\\
\multirow[b]{-5.3}{*}{\includegraphics[height=65px]{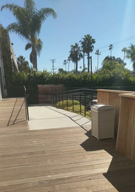}} &
\includegraphics[width=20px]{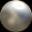} &
\includegraphics[width=20px]{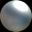}&  
\includegraphics[width=20px]{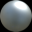}&   
\includegraphics[width=20px]{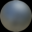}&  
\includegraphics[width=20px]{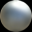}& 
\includegraphics[width=20px]{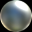}&   
(s)\\
&
\includegraphics[width=20px]{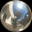} &
\includegraphics[width=20px]{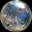}&  
\includegraphics[width=20px]{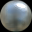}&   
\includegraphics[width=20px]{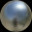}&  
\includegraphics[width=20px]{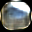}& 
\includegraphics[width=20px]{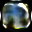}&   
(m)\\[1ex]

\end{tabular}
    \vspace{-10pt}
	\caption{\small Ablation study: Unseen image inputs, ground truth, and rendered images of diffuse(d), matte silver(s) and mirror(m) spheres, lit with HDR lighting inference from networks trained using different loss terms(top). Our full method is labeled in bold.}
	\label{fig:ablate}
\vspace{-15pt}
\end{figure}

\vspace{-8pt} \paragraph{Ablation studies:}
We assess the importance of the different loss terms, $L_\text{rec}$ for each BRDF and $L_\text{adv}$, and report $L_\text{rec}$ and $\theta_\text{rgb}$ for networks supervised using subsets of the loss terms in Table \ref{Table:1}. Training with only the mirror BRDF or only the diffuse BRDF leads to higher $L_\text{rec}$ for the others. However, training with only the matte silver BRDF still yields low $L_\text{rec}$ for the diffuse sphere, suggesting they reveal similar lighting cues. In Fig.\ \ref{fig:ablate}, we show the ground truth images and renderings produced for each loss variant. Visually, training with only the mirror ball $L_{1(m)}$ fails to recover the full dynamic range of lighting, as expected. Training with only the matte silver $L_{1(s)}$ or diffuse $L_{1(d)}$ fails to produce a realistic mirror ball; thus objects with sharp specular reflections could not be plausibly rendered. Training with $L_\text{adv}$ yields higher frequency illumination as expected.

\begin{table}[h]
\caption{\small Average $L_{1}$ loss by BRDF: diffuse (d), mirror (m), and matte silver (s), and RGB angular error $\theta_\text{rgb}$ for diffuse (columns), for our network trained with different loss terms (rows). We compare ground truth images with those \textit{rendered} using our HDR lighting inference, for \textit{unseen} indoor and outdoor locations.}
\vspace{-5pt}
\footnotesize
\centering
\begin{tabular}{@{}l@{\quad}c@{\quad}c@{}c@{\quad\;}c@{\quad}c@{}c@{\quad\;}c@{\quad}c@{}c@{\quad\;}c@{\quad}c@{}}
\toprule
 & \multicolumn{2}{@{}c@{}}{$L_{1(d)}$} & & \multicolumn{2}{@{}c@{}}{$L_{1(s)}$} & & \multicolumn{2}{@{}c@{}}{$L_{1(m)}$} & & \multicolumn{2}{@{}c@{}}{$\theta_{\text{rgb}(d)}^{\circ}$} \\
\cmidrule{2-3} \cmidrule{5-6} \cmidrule{8-9} \cmidrule{11-12} 
Loss terms & UI & UO & & UI & UO & & UI & UO & & UI & UO \\
\midrule
L\textsubscript{1(m,d,s)}
+ L\textsubscript{adv}    & 0.12 & 0.13 & & 0.13 & 0.13 & & 0.17 & 0.16 & &  9.8 & 10.8 \\
L\textsubscript{1(m,d,s)} & 0.12 & 0.13 & & 0.12 & 0.13 & & 0.15 & 0.14 & &  9.9 & 11.0 \\
L\textsubscript{1(m)}     & 0.20 & 0.18 & & 0.16 & 0.15 & & 0.14 & 0.13 & & 11.0 & 13.5 \\
L\textsubscript{1(s)}     & 0.12 & 0.13 & & 0.13 & 0.13 & & 0.21 & 0.20 & & 10.0 & 11.4 \\
L\textsubscript{1(d)}     & 0.12 & 0.13 & & 0.15 & 0.15 & & 0.28 & 0.27 & & 10.0 & 11.2\\
\bottomrule
\end{tabular}
\vspace{-10pt}
\label{Table:1}
\end{table}

\subsection{Qualitative Results}
\begin{figure}[ht]
\centering
\scriptsize
\begin{tabular}{c@{ }c@{ }c@{ }c@{ }c@{\hspace{4pt}}c@{ }c@{ }c@{ }c@{ }l@{ }}
& \scriptsize UI input & \scriptsize (d) & \scriptsize (s) & \scriptsize (m) &
\scriptsize UO input & \scriptsize (d) & \scriptsize (s) & \scriptsize (m) & \\
\multirow[c]{-2}{*}{\scriptsize \rotatebox{90}{25\textsuperscript{th} \%}} & \multirow[c]{-2.8}{*}{\includegraphics[height=42px]{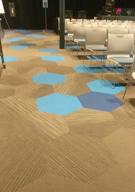}}&
\includegraphics[width=20px]{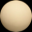}&  
\includegraphics[width=20px]{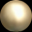}&  
\includegraphics[width=20px]{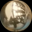}&
\multirow[c]{-2.8}{*}{\includegraphics[height=42px]{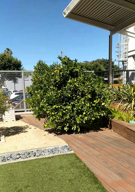}}&  
\includegraphics[width=20px]{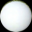}&  
\includegraphics[width=20px]{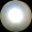}&  
\includegraphics[width=20px]{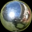}& \scriptsize gt\\
& 
&  
\includegraphics[width=20px]{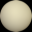}&  
\includegraphics[width=20px]{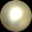}&    
\includegraphics[width=20px]{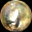}&                     
&  
\includegraphics[width=20px]{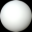}&  
\includegraphics[width=20px]{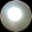}&  
\includegraphics[width=20px]{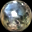}& \scriptsize pred \\[1ex]

\multirow[c]{-2}{*}{\scriptsize \rotatebox{90}{50\textsuperscript{th} \%}} & \multirow[c]{-2.8}{*}{\includegraphics[height=42px]{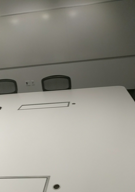}} &
\includegraphics[width=20px]{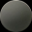} &  
\includegraphics[width=20px]{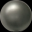} &  
\includegraphics[width=20px]{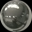} &
\multirow[c]{-2.8}{*}{\includegraphics[height=42px]{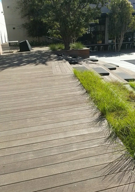}}&  
\includegraphics[width=20px]{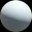} &  
\includegraphics[width=20px]{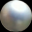} &  
\includegraphics[width=20px]{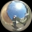} & \scriptsize gt\\
& 
&  
\includegraphics[width=20px]{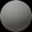} &  
\includegraphics[width=20px]{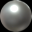} &    
\includegraphics[width=20px]{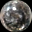} &                     
&  
\includegraphics[width=20px]{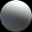} &  
\includegraphics[width=20px]{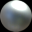} &  
\includegraphics[width=20px]{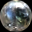} & \scriptsize pred \\[1ex]

\multirow[c]{-2}{*}{\scriptsize \rotatebox{90}{75\textsuperscript{th} \%}} & \multirow[c]{-2.8}{*}{\includegraphics[height=42px]{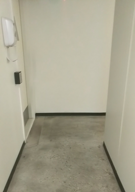}} &
\includegraphics[width=20px]{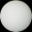} &  
\includegraphics[width=20px]{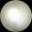} &  
\includegraphics[width=20px]{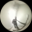} &
\multirow[c]{-2.8}{*}{\includegraphics[height=42px]{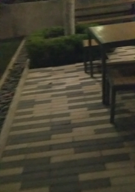}}&  
\includegraphics[width=20px]{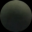} &  
\includegraphics[width=20px]{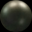} &  
\includegraphics[width=20px]{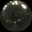} & \scriptsize gt\\
& 
&  
\includegraphics[width=20px]{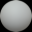} &  
\includegraphics[width=20px]{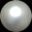} &    
\includegraphics[width=20px]{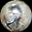} &                     
&  
\includegraphics[width=20px]{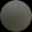} &  
\includegraphics[width=20px]{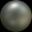} &  
\includegraphics[width=20px]{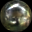} & \scriptsize pred \\[1ex]

\end{tabular}
\vspace{-12pt}
\caption{\small Qualitative comparisons between ground truth spheres and renderings using our HDR lighting inference and IBRL. Examples shown for 25\textsuperscript{th}, 50\textsuperscript{th}, and 75\textsuperscript{th} percentiles for $L_\text{rec}$.}
\label{fig:spheres_25_50_75}
\vspace{-15pt}
\end{figure}

\paragraph{Ground truth comparisons:} 
In Fig.\ \ref{fig:spheres_25_50_75}, we show examples of ground truth spheres compared with those rendered using image-based relighting and our HDR lighting inference, for each BRDF. These examples correspond to the 25\textsuperscript{th}, 50\textsuperscript{th}, and 75\textsuperscript{th} percentiles for the $L_\text{rec}$ loss.

\begin{figure*}[h]
    \scriptsize
    \vspace{-6pt}
	\centerline{
		\begin{tabular}{c@{ }c@{\quad}c@{ }c@{ }c@{ }c@{\quad}c@{ }c@{ }c@{ }c@{ }}
		    & (a) unseen input & (b) real object & (c) GT HDR IBL & (d) ours & (e) \cite{Gardner:2017:Indoor}/\cite{hold:2017:deep}
		          & (f) real object & (g) GT HDR IBL & (h) ours & (i) \cite{Gardner:2017:Indoor}/\cite{hold:2017:deep}\\
		    \rotatebox{90}{\footnotesize indoor} &
			\includegraphics[width=0.7in]{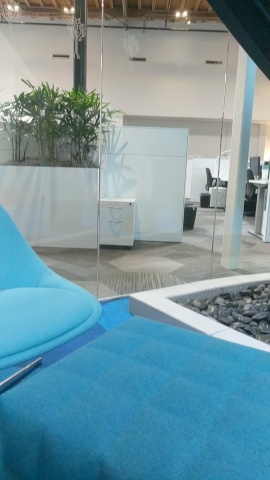} &
			\includegraphics[width=0.7in]{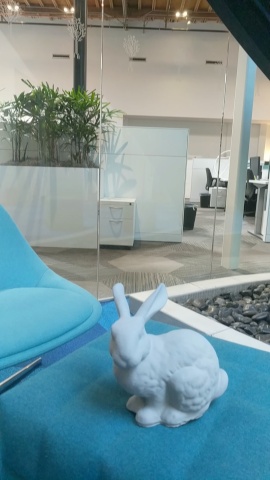} &
			\includegraphics[width=0.7in]{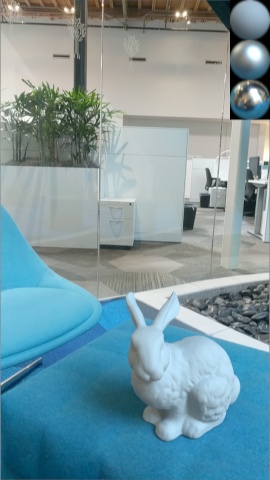} &
			\includegraphics[width=0.7in]{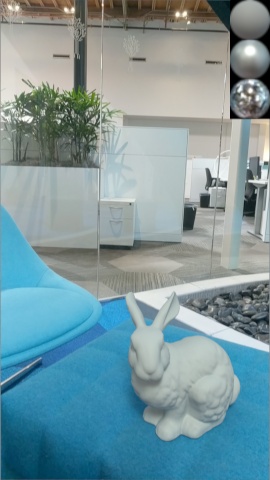} &
			\includegraphics[width=0.7in]{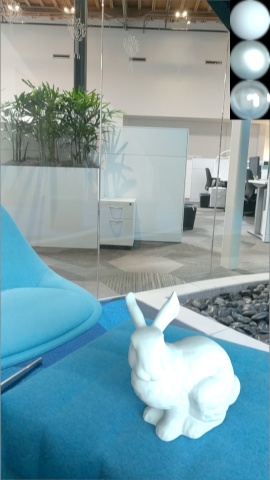} &
			\includegraphics[width=0.7in]{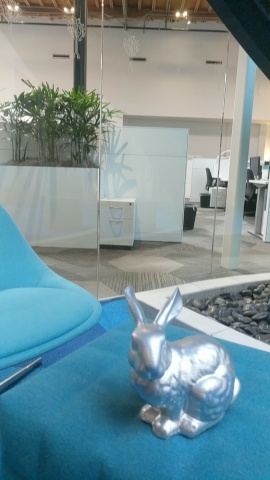} &
			\includegraphics[width=0.7in]{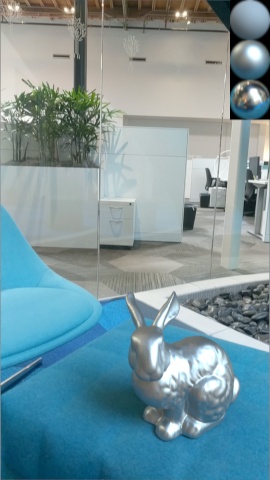} &
			\includegraphics[width=0.7in]{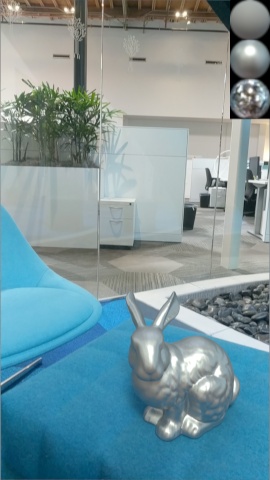} &
			\includegraphics[width=0.7in]{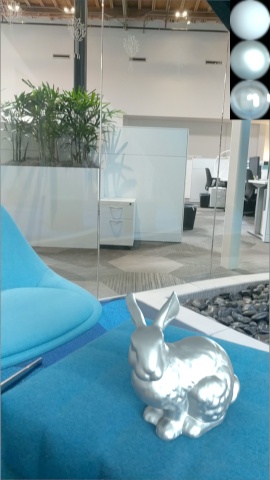} \\
			\rotatebox{90}{\footnotesize outdoor shade} &
            \includegraphics[width=0.7in]{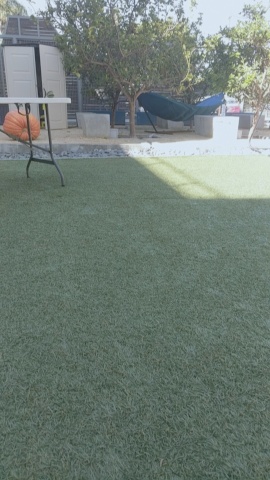} &
			\includegraphics[width=0.7in]{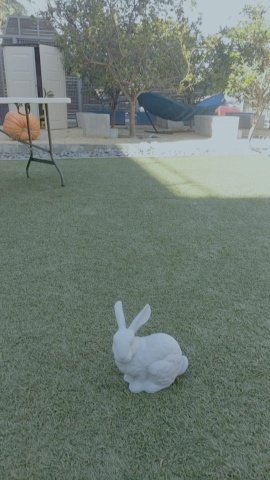} &
			\includegraphics[width=0.7in]{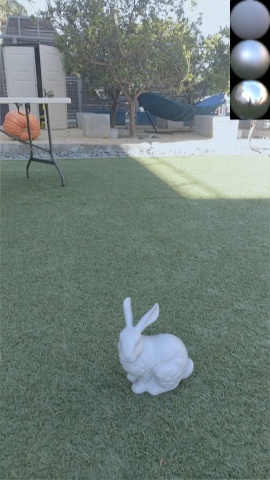} &
			\includegraphics[width=0.7in]{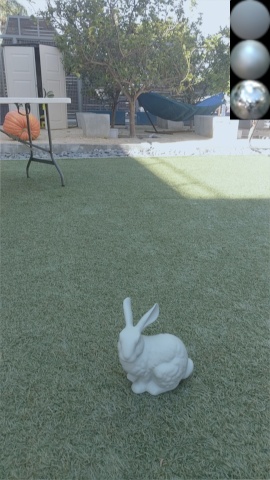} &
			\includegraphics[width=0.7in]{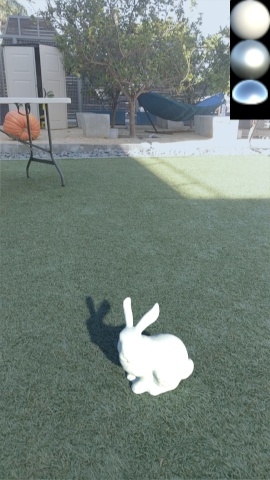} &
			\includegraphics[width=0.7in]{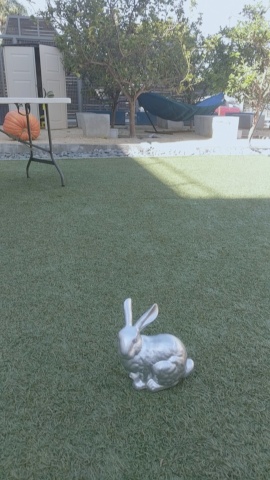} &
			\includegraphics[width=0.7in]{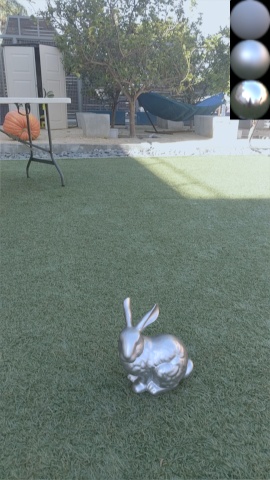} &
			\includegraphics[width=0.7in]{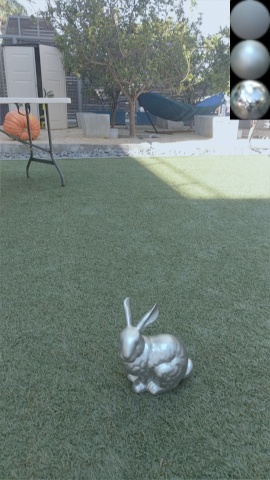} &
			\includegraphics[width=0.7in]{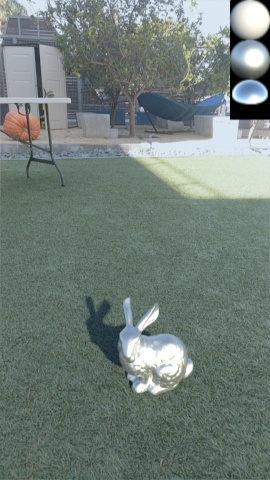} \\
			\rotatebox{90}{\footnotesize outdoor sun} &
			\includegraphics[width=0.7in]{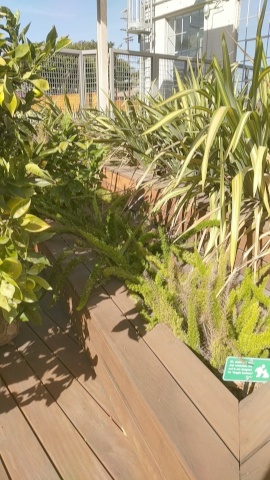} &
			\includegraphics[width=0.7in]{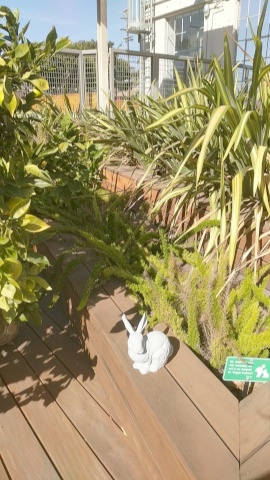} &
			\includegraphics[width=0.7in]{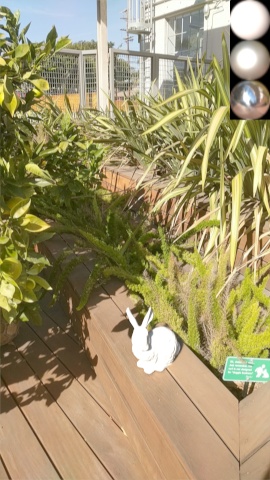} &
			\includegraphics[width=0.7in]{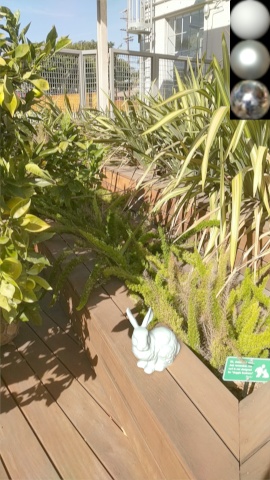} &
			\includegraphics[width=0.7in]{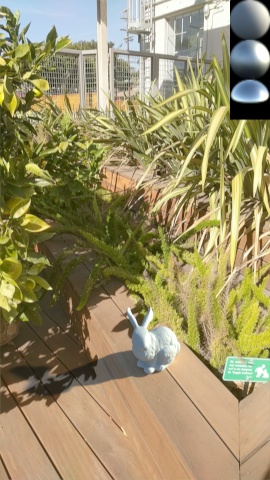} &
			\includegraphics[width=0.7in]{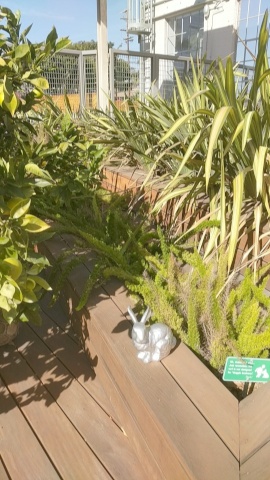} &
			\includegraphics[width=0.7in]{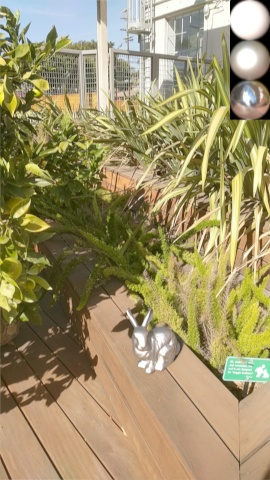} &
			\includegraphics[width=0.7in]{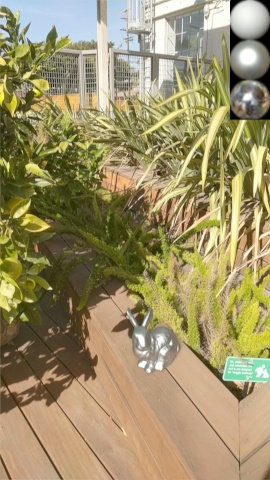} &
			\includegraphics[width=0.7in]{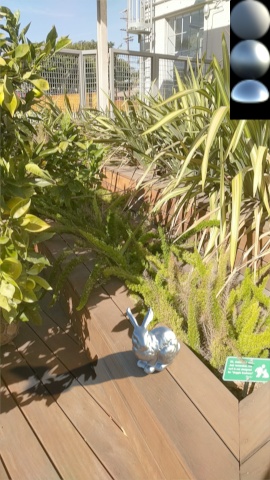} \\
		\end{tabular}}
		\vspace{-5pt}
		\caption{\small For each input image (a), we photograph a real 3D-printed bunny placed in the scene for two different BRDFs (b, f) and capture ground truth HDR panoramas at the bunny's location. Using GI rendering with IBL, we render a virtual bunny into the scene using ground truth lighting (c, g), our lighting inference (d, h), and that of the state-of-the-art methods for indoor \cite{Gardner:2017:Indoor} or outdoor \cite{hold:2017:deep} scenes (e, i).}
		\vspace{-10pt}
		\label{fig:main_bunnies}
\end{figure*}

\vspace{-5pt} \paragraph{Virtual object relighting:}
We 3D-print two identical bunnies using the model from \cite{Stanford3DScanningRepository}. The two are coated with paints of measured reflectance: diffuse gray (34.5\% reflective) and matte silver (49.9\% reflective), respectively. We photograph these ``real" bunnies in different scenes using the Google Pixel XL, also capturing a clean plate for lighting inference and virtual object compositing. In Fig.\ \ref{fig:main_bunnies} we compare the real bunny images (b, f) to off-line rendered composites using our lighting estimates (d, h) (IBL rendering described in the supplemental materials). We also record ground truth HDR lighting as in \cite{debevec:1998:rendering} using a Canon 5D Mark III, color correcting the raw linear HDR panorama so it matches the LDR phone image. We fit a linearization curve for each LDR input using a color chart, however the phone's image-dependent tone-mapping makes radiometric alignment challenging. We compare renderings using the ground truth and predicted lighting in Fig.\ \ref{fig:main_bunnies} (c, g).

\subsection{Comparisons with Previous Work}
\begin{figure}[h]
	\centerline{
		\begin{tabular}{@{ }c@{ }} 
			\includegraphics[width=2.8in]{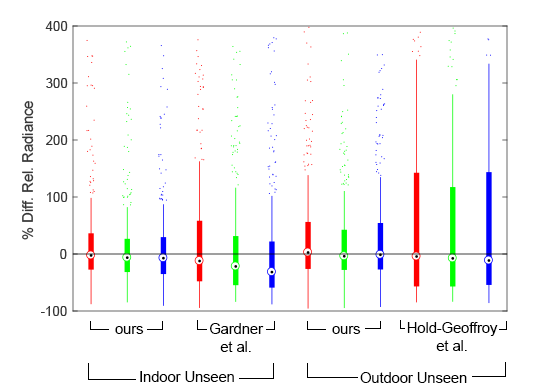} \\
		\end{tabular}}
		\vspace{-10pt}
		\caption{\small Boxplot of RGB relative radiance accuracy, measured by summing linear pixel values of the diffuse ball rendered with the HDR lighting estimates, and comparing with ground truth: (pred-gt)/gt, $n=450$, for our approach and the previous state-of-the-art methods for indoor\cite{Gardner:2017:Indoor} and outdoor\cite{hold:2017:deep} scenes.}
		\label{fig:lalonde_radiance}
		\vspace{-1pt}
\end{figure}

We retrain our network for the 3:4 aspect ratio input of the state-of-the-art methods for indoor \cite{Gardner:2017:Indoor} and outdoor \cite{hold:2017:deep} scenes, cropping a $1080 \times 810$ landscape image from the center of each portrait input and resizing to $192 \times 144$ to maintain our FC layer size. (Our comparison network thus observes half of the FOV of our standard network.) Gardner et al. \cite{Gardner:2017:Indoor} host a server to predict HDR lighting given an input image; Hold-Geoffroy et al. \cite{hold:2017:deep} also predict camera elevation. We randomly select 450 images from test sets UI and UO and retrieve their lighting estimates as HDR panoramas, converting them to the $32\times32$ mirror ball mapping and rotating them to camera space using the predicted camera elevation if given. We render spheres of each BRDF with IBRL and compare with ground truth, showing the average $L_{1}$ loss for each BRDF and $\theta_\text{rgb}$ for the diffuse ball in Table \ref{table:lalonde}. We also show the relative error in total scene radiance measured by summing all diffuse sphere linear pixel values\footnote{Scene radiance is modulated by the albedo and foreshortening factor of the diffuse sphere, with greater frontal support, and we use $\gamma=2.2$.} in Table \ref{fig:lalonde_radiance}. We show comparison sphere renderings in Fig.\ \ref{fig:comparisons_qual} and bunny renderings in Fig.\ \ref{fig:main_bunnies} (e, i), with more in supplemental materials along with a perceptual user study. We show significant improvements compared to both approaches, while requiring only one model that generalizes to both indoor and outdoor scenes. Without a specific sun and sky model, our network also infers diverse light sources for outdoor scenes. However, we present these results with two caveats: first, our training data are generated with a fixed FOV camera, which was varied and unknown for previous approaches, and second, our training and test data are generated with the same camera. Nonetheless, for mobile mixed-reality with a fixed-FOV, we show that optimizing for accurately rendered objects for multiple BRDFs improves lighting estimation.

\begin{table}[]
\caption{\small Quantitative comparisons with the previous state-of-the-art in indoor\cite{Gardner:2017:Indoor} and outdoor\cite{hold:2017:deep} lighting estimation. Average $L_{1}$ loss by BRDF: diffuse(d), mirror(m), and matte silver(s), and RGB angular error $\theta_\text{rgb}$ for the diffuse sphere. $n=450$ for each.}
\vspace{-5pt}
\footnotesize
\centering
\begin{tabular}{@{}l@{\quad}r@{}c@{}l@{}c@{\quad\;}r@{}c@{}l@{}c@{\quad\;}r@{}c@{}l@{}c@{\quad\;}r@{}c@{}l@{}}
\toprule
  & \multicolumn{7}{@{}c@{}}{unseen indoor (UI)} &
  & \multicolumn{7}{@{}c@{}}{unseen outdoor (UO)} \\
\cmidrule{2-8}\cmidrule{10-16}
  & \multicolumn{3}{@{}c@{}}{ours} & & \multicolumn{3}{@{}c@{}}{\cite{Gardner:2017:Indoor}} &
  & \multicolumn{3}{@{}c@{}}{ours} & & \multicolumn{3}{@{}c@{}}{\cite{hold:2017:deep}} \\
\midrule
$L_{1(d)}$ &
    $\mathbf{0.13}$ & $\pm$ & $\mathbf{0.07}$ & & $0.21$ & $\pm$ & $0.11$ & &
    $\mathbf{0.13}$ & $\pm$ & $\mathbf{0.08}$ & & $0.25$ & $\pm$ & $0.12$ \\
$L_{1(s)}$ &
    $\mathbf{0.14}$ & $\pm$ & $\mathbf{0.05}$ & & $0.22$ & $\pm$ & $0.06$ & &
    $\mathbf{0.14}$ & $\pm$ & $\mathbf{0.06}$ & & $0.25$ & $\pm$ & $0.07$ \\
$L_{1(m)}$ &
    $\mathbf{0.18}$ & $\pm$ & $\mathbf{0.03}$ & & $0.23$ & $\pm$ & $0.06$ & &
    $\mathbf{0.17}$ & $\pm$ & $\mathbf{0.04}$ & & $0.34$ & $\pm$ & $0.06$ \\
$\theta_{\text{rgb}(d)}^{\circ}$ &
    $\mathbf{10.3}$ & $\pm$ & $\mathbf{8.8^{\circ}}$  & & $11.9$ & $\pm$ & $7.2^{\circ}$ & &
    $\mathbf{11.2}$ & $\pm$ & $\mathbf{10.9^{\circ}}$ & & $14.3$ & $\pm$ & $6.6^{\circ}$ \\
\bottomrule
\end{tabular}
\vspace{-15pt}
\label{table:lalonde}
\end{table}

\begin{figure}[t]
\centering
\scriptsize
\begin{tabular}{@{}c@{ }c@{ }c@{ }c@{\hspace{3pt}}|@{\hspace{3pt}}c@{ }c@{ }c@{ }c@{ }l@{ }}
\scriptsize UI input & \scriptsize ours & \scriptsize gt & \scriptsize \cite{Gardner:2017:Indoor} &
\scriptsize UO input & \scriptsize ours & \scriptsize gt & \scriptsize \cite{hold:2017:deep} & \\
\hline\\[-2.0ex]
\multirow[c]{-0.5}{*}{\includegraphics[height=30px]{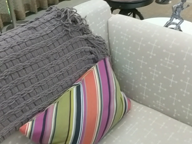}}&
\includegraphics[width=20px]{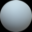}&  
\includegraphics[width=20px]{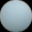}&  
\includegraphics[width=20px]{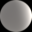}&
\multirow[c]{-0.5}{*}{\includegraphics[height=30px]{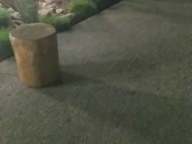}}&  
\includegraphics[width=20px]{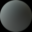}&  
\includegraphics[width=20px]{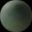}&  
\includegraphics[width=20px]{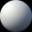}& \scriptsize (d)\\
&  
\includegraphics[width=20px]{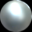}&  
\includegraphics[width=20px]{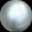}&   
\includegraphics[width=20px]{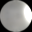}& 
&  
\includegraphics[width=20px]{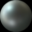}&  
\includegraphics[width=20px]{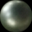}&  
\includegraphics[width=20px]{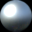}& \scriptsize (s)\\
& 
\includegraphics[width=20px]{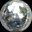}&  
\includegraphics[width=20px]{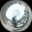}&    
\includegraphics[width=20px]{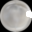}&      
&  
\includegraphics[width=20px]{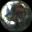}&  
\includegraphics[width=20px]{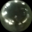}&  
\includegraphics[width=20px]{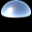}& \scriptsize (m)\\
\hline\\[-2.0ex]
\multirow[c]{-0.5}{*}{\includegraphics[height=30px]{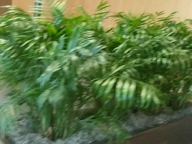}}&
\includegraphics[width=20px]{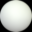}&  
\includegraphics[width=20px]{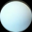}&  
\includegraphics[width=20px]{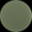}&
\multirow[c]{-0.5}{*}{\includegraphics[height=30px]{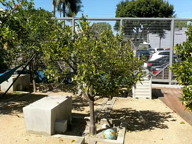}}&  
\includegraphics[width=20px]{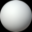}&  
\includegraphics[width=20px]{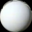}&  
\includegraphics[width=20px]{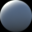}& \scriptsize (d)\\
&  
\includegraphics[width=20px]{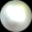}&  
\includegraphics[width=20px]{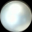}&   
\includegraphics[width=20px]{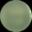}& 
&  
\includegraphics[width=20px]{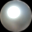}&  
\includegraphics[width=20px]{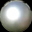}&  
\includegraphics[width=20px]{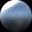}& \scriptsize (s)\\
& 
\includegraphics[width=20px]{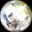}&  
\includegraphics[width=20px]{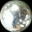}&    
\includegraphics[width=20px]{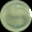}&      
&  
\includegraphics[width=20px]{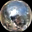}&  
\includegraphics[width=20px]{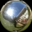}&  
\includegraphics[width=20px]{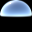}& \scriptsize (m)\\
\hline\\[-2.0ex]
\multirow[c]{-0.5}{*}{\includegraphics[height=30px]{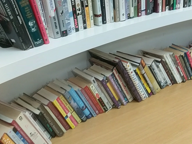}}&
\includegraphics[width=20px]{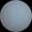}&  
\includegraphics[width=20px]{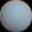}&  
\includegraphics[width=20px]{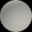}&
\multirow[c]{-0.5}{*}{\includegraphics[height=30px]{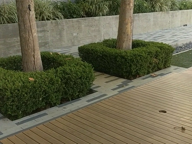}}&  
\includegraphics[width=20px]{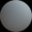}&  
\includegraphics[width=20px]{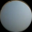}&  
\includegraphics[width=20px]{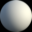}& \scriptsize (d)\\
&  
\includegraphics[width=20px]{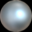}&  
\includegraphics[width=20px]{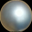}&   
\includegraphics[width=20px]{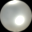}& 
&  
\includegraphics[width=20px]{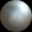}&  
\includegraphics[width=20px]{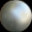}&  
\includegraphics[width=20px]{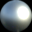}& \scriptsize (s)\\
& 
\includegraphics[width=20px]{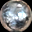}&  
\includegraphics[width=20px]{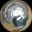}&    
\includegraphics[width=20px]{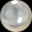}&      
&  
\includegraphics[width=20px]{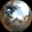}&  
\includegraphics[width=20px]{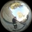}&  
\includegraphics[width=20px]{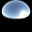}& \scriptsize (m)\\[1ex]
\end{tabular}
\vspace{-10pt}
\caption{\small Ground truth and rendered spheres produced via IBRL using our predicted HDR lighting and that of the previous state-of-the-art for indoor \cite{Gardner:2017:Indoor} and outdoor \cite{hold:2017:deep} scenes.}
\label{fig:comparisons_qual}
\vspace{-15pt}
\end{figure}

\begin{figure}[ht]
\centering
\scriptsize
\vspace{-5pt}
\begin{tabular}{@{}c@{}c@{}c@{}c@{ }c@{}c@{}c@{}c@{ }c@{}c@{}c@{}c@{ }l@{}}
\multicolumn{3}{@{}c@{}}{frame 0} &
\multicolumn{3}{@{}c@{}}{frame 1} &
\multicolumn{3}{@{}c@{}}{frame 2} &
\multicolumn{3}{@{}c@{}}{frame 3} &\\
\multicolumn{3}{@{}c@{}}{ \includegraphics[width=54.5px]{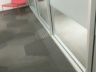} } &
\multicolumn{3}{@{}c@{}}{ \includegraphics[width=54.5px]{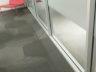} } &
\multicolumn{3}{@{}c@{}}{ \includegraphics[width=54.5px]{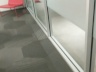} } &
\multicolumn{3}{@{}c@{}}{ \includegraphics[width=54.5px]{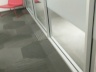} } &\\
0 & 1 & 2 & 3 & 0 & 1 & 2 & 3 & 0 & 1 & 2 & 3 &\\
\includegraphics[width=18.5px]{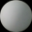} &
\includegraphics[width=18.5px]{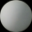} &
\includegraphics[width=18.5px]{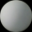} &
\includegraphics[width=18.5px]{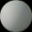} &
\includegraphics[width=18.5px]{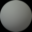} &
\includegraphics[width=18.5px]{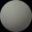} &
\includegraphics[width=18.5px]{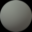} &
\includegraphics[width=18.5px]{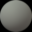} &
\includegraphics[width=18.5px]{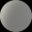} &
\includegraphics[width=18.5px]{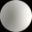} &
\includegraphics[width=18.5px]{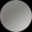} &
\includegraphics[width=18.5px]{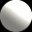} & 
(d)\\
\includegraphics[width=18.5px]{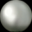} &
\includegraphics[width=18.5px]{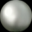} &
\includegraphics[width=18.5px]{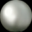} &
\includegraphics[width=18.5px]{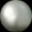} &
\includegraphics[width=18.5px]{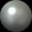} &
\includegraphics[width=18.5px]{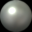} &
\includegraphics[width=18.5px]{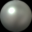} &
\includegraphics[width=18.5px]{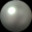} &
\includegraphics[width=18.5px]{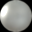} &
\includegraphics[width=18.5px]{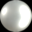} &
\includegraphics[width=18.5px]{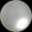} &
\includegraphics[width=18.5px]{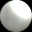} & 
(s)\\
\includegraphics[width=18.5px]{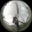} &
\includegraphics[width=18.5px]{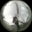} &
\includegraphics[width=18.5px]{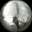} &
\includegraphics[width=18.5px]{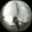} &
\includegraphics[width=18.5px]{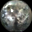} &
\includegraphics[width=18.5px]{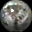} &
\includegraphics[width=18.5px]{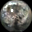} &
\includegraphics[width=18.5px]{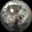} &
\includegraphics[width=18.5px]{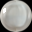} &
\includegraphics[width=18.5px]{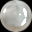} &
\includegraphics[width=18.5px]{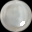} &
\includegraphics[width=18.5px]{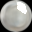} & 
(m)\\
\multicolumn{4}{@{}c@{}}{(a) ground truth} &
\multicolumn{4}{@{}c@{}}{(b) ours} & 
\multicolumn{4}{@{}c@{}}{(c) \cite{Gardner:2017:Indoor}} & \\
\end{tabular}
\vspace{-8pt}
	\caption{\small Example ground truth spheres (a) and renderings produced with IBRL using our predicted illumination (b) and that of \cite{Gardner:2017:Indoor} (c), for four sequential UI video frames (top).}
	\label{fig:time}
\vspace{-12pt}
\end{figure}

\vspace{-10pt}\paragraph{Temporal consistency:}
We do not explicitly optimize for temporal consistency, but the adjacent video frames in our training data provide an indirect form of temporal regularization. In Fig.\ \ref{fig:time} we compare rendered results from four sequential frames for our approach and for that of Gardner et al. \cite{Gardner:2017:Indoor}. While we show qualitative improvement, adding a temporal loss term is of interest for future work.

\subsection{Performance and Demonstration}
Our inference runs at 12-20 fps on various mobile phone CPUs. We report performance for smaller networks and output lighting resolutions and timing for specific mobile phones in supplemental materials. We also authored a demo mobile application to predict lighting and render plausibly-lit virtual objects at interactive frame rates, using real-time pre-computed radiance transfer \cite{Sloan:2002:PRT} rendering.

\section{Limitations and Future Work}
\paragraph{Spatially-varying illumination:}
The reference spheres of the training data reflect the illumination from a point 60 cm in front of the camera and do not reveal spatially-varying lighting cues.  Virtual AR objects are often placed on surfaces visible in the scene, and the light bouncing up from the surface should be the illumination on the object coming from below.  A potential improvement to our technique would be to replace the bottom directions of our lighting estimate with pixel values sampled from the scene surface below each object, allowing objects placed in different parts of the scene to receive differently colored bounce light from their environments.

\begin{figure}[h]
\centering
\scriptsize
\vspace{-5pt}
\begin{tabular}{c@{ }c@{ }c@{ }c@{\hspace{7pt}}c@{ }c@{ }c@{ }c@{ }l@{ }}
\scriptsize SI input & \scriptsize (d) & \scriptsize (s) & \scriptsize (m) &
\scriptsize SO input & \scriptsize (d) & \scriptsize (s) & \scriptsize (m) & \\

\multirow[c]{-2.75}{*}{\includegraphics[height=42px]{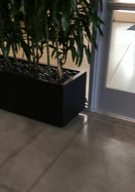}}&
\includegraphics[width=20px]{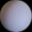}&  
\includegraphics[width=20px]{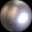}&  
\includegraphics[width=20px]{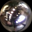}&
\multirow[c]{-2.75}{*}{\includegraphics[height=42px]{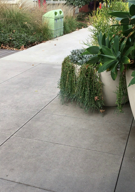}}&  
\includegraphics[width=20px]{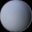}&  
\includegraphics[width=20px]{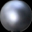}&  
\includegraphics[width=20px]{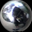}& \scriptsize gt\\
&  
\includegraphics[width=20px]{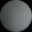}&  
\includegraphics[width=20px]{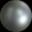}&    
\includegraphics[width=20px]{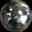}&                     
&  
\includegraphics[width=20px]{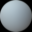}&  
\includegraphics[width=20px]{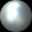}&  
\includegraphics[width=20px]{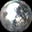}& \scriptsize pred \\[1ex]

\end{tabular}
\vspace{-8pt}
\caption{\small Example ground  truth  spheres  and  renderings  produced  with  IBRL  using  our  predicted HDR lighting, with input images from a different camera.}
\vspace{-15pt}
\label{fig:other_camera}
\end{figure}

\paragraph{Using a different camera:}
Our test and training data are captured with the same camera. In Fig.\ \ref{fig:other_camera} we show results for two images captured using a different mobile phone camera (Apple iPhone 6). Qualitatively, we observe differences in white balance, suggesting an avenue for future work. Similarly, our network is trained for a particular camera FOV and may not generalize to others. 

\begin{figure}[h]
\centering
\scriptsize
\vspace{-5pt}
\begin{tabular}{c@{ }c@{ }c@{ }c@{\hspace{7pt}}c@{ }c@{ }c@{ }c@{ }l@{ }}
\scriptsize UI input & \scriptsize (d) & \scriptsize (s) & \scriptsize (m) &
\scriptsize UI input & \scriptsize (d) & \scriptsize (s) & \scriptsize (m) & \\

\multirow[c]{-2.75}{*}{\includegraphics[height=42px]{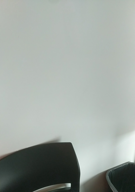}}&
\includegraphics[width=20px]{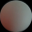}&  
\includegraphics[width=20px]{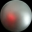}&  
\includegraphics[width=20px]{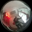}&
\multirow[c]{-2.75}{*}{\includegraphics[height=42px]{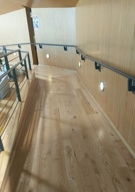}}&  
\includegraphics[width=20px]{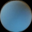}&  
\includegraphics[width=20px]{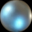}&  
\includegraphics[width=20px]{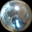}& \scriptsize gt\\
&  
\includegraphics[width=20px]{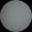}&  
\includegraphics[width=20px]{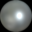}&    
\includegraphics[width=20px]{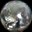}&                     
&  
\includegraphics[width=20px]{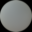}&  
\includegraphics[width=20px]{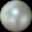}&  
\includegraphics[width=20px]{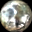}& \scriptsize pred \\[1ex]

\end{tabular}
\vspace{-10pt}
\caption{\small Example challenging scenes: ground  truth  spheres  and  renderings  produced  with  IBRL  using  our  predicted HDR lighting.}
\vspace{-15pt}
\label{fig:limitations}
\end{figure}

\paragraph{Challenging image content:}
Simple scenes lacking variation in surface normals and albedo (Fig.\ \ref{fig:limitations}, left) can challenge our inference approach, and scenes dominated by a strongly hued material can also pose a challenge (Fig.\ \ref{fig:limitations}, right).  Adding knowledge of the camera's exposure and white balance used for each input image might improve the robustness of the inference.

\paragraph{Future work:}
During mobile mixed reality sessions, objects are positioned on planes detected using sensor data fused with structure-from-motion \cite{Ullman:1979:sfm}. Thus, computational resources are already devoted to geometric reasoning, which would be of interest to leverage for improved mixed reality lighting estimation. Furthermore, inertial measurements could be leveraged to continuously fuse and update lighting estimates as a user moves a phone throughout an environment. Similarly, as our training data already includes temporal structure, explicitly optimizing for temporal stability would be of interest. Lastly, one could increase generality by acquiring training data in a raw video format and simulating different camera models during training.

\section{Conclusion}
We have presented an HDR lighting inference method for mobile mixed reality, trained using only LDR imagery, leveraging reference spheres with different materials to reveal different lighting cues in a single exposure. This work is the first CNN-based approach that generalizes to both indoor and outdoor scenes for a single input image, with improved lighting estimation for mobile mixed reality as compared to previous work developed to handle only a single class of lighting.

{\small
\bibliographystyle{ieee}
\bibliography{lighting_estimation}
}

\end{document}